\documentclass[runningheads]{llncs}
\usepackage{graphicx}

\usepackage{amsmath}
\usepackage{amssymb}
\usepackage{siunitx}
\usepackage{glossaries}
\usepackage{subfig}
\usepackage[noend]{algorithm2e}
\DontPrintSemicolon
\usepackage[misc]{ifsym}

\newacronym{HTM}{HTM}{Hierarchical Temporal Memory}
\newacronym[plural=SDRs,longplural=Sparse Distributed Representations]{SDR}{SDR}{Sparse Distributed Representation}
\newacronym{RL}{RL}{Reinforcement Learning}
\newacronym{ML}{ML}{Machine Learning}
\newacronym{MDP}{MDP}{Markov Decision Process}
\newacronym{TD}{TD}{Temporal Difference}
\newacronym{MPF}{MPF}{Memory-Prediction Framework}
\newacronym{HSAT}{HSAT}{Hebbian Synapses with Adaptive Thresholds}
\newacronym[longplural=Spiking Neural Networks]{SNN}{SNN}{Spiking Neural Network}
\newacronym{RNN}{RNN}{Recurrent Neural Network}
\newacronym{DQN}{DQN}{Deep Q-Learning}
\newacronym{DRL}{DRL}{Deep Reinforcement Learning}
\newacronym{Meta-RL}{Meta-RL}{Meta-Reinforcement Learning}
\newacronym{MAML}{MAML}{Model-Agnostic Meta-Learning}

\newcommand\myeq{\mkern1.5mu{=}\mkern1.5mu}

\begin{document}
\title{HTMRL: Biologically Plausible Reinforcement Learning with Hierarchical Temporal Memory\thanks{Supported by the Research Foundation - Flanders (fwo): PhD Fellowship 1SB0719N.}}
\titlerunning{Reinforcement Learning with Hierarchical Temporal Memory}
%
\author{Jakob Struye \orcidID{0000-0003-1360-7672} \and
Kevin Mets\orcidID{0000-0002-4812-4841} \and
Steven Latr\'e\orcidID{0000-0003-0351-1714}} 
\authorrunning{J. Struye et al.}
%
\institute{University of Antwerp - imec\\
IDLab - Department of Mathematics and Computer Science\\
Sint-Pietersvliet 7, 2000 Antwerp, Belgium\\
\email{firstname.lastname@uantwerpen.be}}
\maketitle              
\begin{abstract}
Building Reinforcement Learning (RL) algorithms which are able to adapt to continuously evolving tasks is an open research challenge. One technology that is known to inherently handle such non-stationary input patterns well is Hierarchical Temporal Memory (HTM), a general and biologically plausible computational model for the human neocortex. As the RL paradigm is inspired by human learning, HTM is a natural framework for an RL algorithm supporting non-stationary environments. In this paper, we present HTMRL, the first strictly HTM-based RL algorithm. We empirically and statistically show that HTMRL scales to many states and actions, and demonstrate that HTM's ability for adapting to changing patterns extends to RL. Specifically, HTMRL performs well on a 10-armed bandit after 750 steps, but only needs a third of that to adapt to the bandit suddenly shuffling its arms. HTMRL is the first iteration of a novel RL approach, with the potential of extending to a capable algorithm for Meta-RL.

\end{abstract}
\section{Introduction}
\gls{RL} is, along with supervised and unsupervised learning, one of the three main pillars of \gls{ML}~\cite{sutton18}. \gls{RL} aims to derive a \textit{policy}, instructing an \textit{agent} on which \textit{action} to take given the current \textit{state} of the \textit{environment}, as to maximise some \textit{reward} signal. Through experience, such policies are gradually improved. The \gls{RL} paradigm evokes human learning: after an initial phase of trial-and-error, one learns to favour behaviour that has previously shown to lead to positive results, while avoiding behaviour expected to produce a negative effect. Since the introduction of \gls{DQN}~\cite{Mnih2013Playing}, the field of \gls{RL} has been dominated by \gls{DRL}, where agents are implemented using deep learning architectures such as neural networks. Although these approaches have greatly increased the scope of problems \gls{RL} can tackle, fundamental issues still remain. One such issue is the inability to adapt in non-stationary environments. For example, a minor perturbation of a video game's background image, not even observable by humans, may cripple a \gls{DRL} agent's performance~\cite{Huang2017Adversarial}. Several \gls{Meta-RL} approaches, such as \gls{MAML}~\cite{Finn2017Model} mitigate the issue by training with many variations of an environment, and encouraging the agent to change quickly when the environment changes.

 We instead propose an \gls{RL} algorithm based on \gls{HTM}, a machine intelligence framework, which has been shown to inherently adapt to changes in input patterns~\cite{Cui2016Continuous,Struye2019Hierarchical}. \gls{HTM} is a model of the neocortex, the most advanced part of the brain, unique to mammals~\cite{hawkins2016bami,hawkins2004intelligence}. This model is not just biologically inspired, but more strongly \textit{biologically constrained}, meaning that concepts that could not plausibly exist in the human neocortex are kept out of \gls{HTM} theory. As no working \gls{RL} algorithms solely using \gls{HTM} currently exist, we first design and evaluate such an algorithm, called HTMRL. Our \textbf{main contributions} are thus (a) proving that \gls{HTM} is a viable approach to \gls{RL}, strengthening the value of the \gls{HTM} framework, often presented as a general framework for intelligence and (b) showing that our HTMRL algorithm behaves well in non-stationary environments, making it a promising approach for \gls{Meta-RL}. The full implementation, along with all experiments, is publicly available\footnote{\url{https://github.com/JakobStruye/HTMRL}}.
 
 The remainder of this paper is structured as follows. Section \ref{sec:bg} provides background on \gls{RL} and \gls{HTM}. In Section \ref{sec:htmrl}, we explain our HTMRL algorithm. Section \ref{sec:results} evaluates HTMRL's capabilities experimentally and statistically, and Section \ref{sec:rw} compares it to related work. Finally, Section \ref{sec:conclusions} summarises our conclusions.
\section{Background} \label{sec:bg}
Both \gls{RL} and \gls{HTM}, the two foundations of this work, are well-established concepts. This section provides a brief overview of the two.
\subsection{Reinforcement Learning}
In \gls{RL}, an agent learns how to behave in an environment~\cite{sutton18}. The entire system is usually modelled as a \gls{MDP}: the 4-tuple
\begin{equation}
(S,A,P,R)
\end{equation}
where $S$ is the set of states ($s \in S$), A is the set of permissible actions ($a \in A$), $P$ is the transition probability function s.t. $P(s_{t+1}|s_t,a_t)$ gives the probability of transitioning to state $s_{t+1}$ after taking action $a_t$ in state $s_t$, and $R$ defines the immediate reward $r_t = R(s_{t+1},s_t,a_t)$ of said transitions. Often, a discount factor $\gamma$ ($\in [0,1]$) is added to the \gls{MDP} as a fifth element, defining how quickly a reward signal should fade away over time. The main goal of a reinforcement learner is then to define a policy $\pi$, mapping observed states to actions. These policies are often probabilistic, meaning actions are sampled given a distribution $P(a|s)$. Ideally, an agent following this policy in an $n$-step environment should attain a maximal discounted sum of rewards, called the \textit{return}, defined as
\begin{equation}
R = \sum^{n}_{t=0}{\gamma^{t}{r_{t}}}
\end{equation}

\subsection{Hierarchical Temporal Memory}
\gls{HTM} is a computational model of the human neocortex, originally proposed by Jeff Hawkins~\cite{hawkins2016bami}. One of the core aspects of \gls{HTM} is its strict biological plausibility, meaning that any feature that could not plausibly be implemented in the neocortex will not be allowed in the \gls{HTM} model. All data within the model is encoded as \glspl{SDR}, binary data structures with only a limited number of \textit{enabled bits} (i.e., with value 1). This mimics how only a small portion of cells in the brain is active at any time. These \glspl{SDR} are generated by \gls{SDR} encoders, responsible for ensuring that partially overlapping enabled bits of two \glspl{SDR} imply that the data points they encode are similar. The \gls{SDR} is then passed to the two main components of the \gls{HTM}: the \textit{spatial pooler} and \textit{temporal memory}.

The spatial pooler reorganises incoming \glspl{SDR} to produce outputs of always the same size and sparsity, and to ensure the available capacity is optimally utilised, offloading these responsibilities from the \gls{SDR} encoders~\cite{Cui2017HTM}. The spatial pooler connects input bits to 2048 output cells 
 through \textit{synapses}. All synapses are created at initialisation, for a portion of bit-cell pairs, but can grow and shrink through time. Each synapse has a (randomly initialised) \textit{permanence} value indicating how strong it is, and only once this permanence reaches a certain threshold does the synapse become \textit{connected}, meaning it can carry a signal from its active input bit to its cell. On each input, the number of incoming signals for each cell is counted, and the 40 cells with the highest count become active. These cells then represent an \gls{SDR} with 40 out of 2048 bits active\footnote{The spatial pooler output size is fixed, as it represents brain tissue, where new cells cannot be grown as needed. By then also fixing the number of active bits, the output has a fixed sparsity, known to improve reliability and robustness of pattern recognition~\cite{Ahmad2016How}. 2048 and 40 were empirically found to perform well, and are widely used across the \gls{HTM} community.}. Synapses are then strengthened if both their input bit and cell are active, but weakened if only the input bit is. This learning algorithm is commonly known as \textit{Hebbian learning}, and the ability of synapses to grow or shrink in response to their endpoints' activations is called \textit{synaptic plasticity}. Algorithm \ref{alg:sp} provides an overview of the spatial pooler's workings.

The next major component is the temporal memory, which expands the spatial pooler's output to also represent the \textit{context} (i.e., previous inputs) in which the input was seen~\cite{Cui2016Continuous}. For example, when the \gls{HTM} is continuously fed the sequence A-C-B-C, the spatial pooler output will eventually be equal for every C, while that of the temporal pooler will depend on whether the C was preceded by an A or by a B. As we only rely on the spatial pooler for the remainder of this paper, we do not go into more detail on temporal memory.
\begin{algorithm}[t]
    \SetKwInput{Precondition}{Precondition~}
    \SetKwInput{Input}{Input~}
    \SetKwInput{State}{State~}
    \SetKwFunction{Encode}{Encode}
    \SetKwFunction{GetCellsWithMostSignals}{GetCellsWithMostSignals}
    \SetKwFunction{Grow}{Grow}
    \SetKwFunction{Shrink}{Shrink}
    \SetKwFunction{IsConnected}{IsConnected}
    \SetKwFunction{EnabledBits}{EnabledBits}
    \SetKwFunction{IncrementSignalCount}{IncrementSignalCount}

    \Input{~State/observation/...}
    \Precondition {~Input matches format expected by encoder.}
    \State{Synapses from input bits to cells}
    \KwOut{40 active cells}
    $enc \gets$ \Encode{input} \tcp{an \gls{SDR}}
    initialise signal counts per cell to 0\;
    \ForEach{$syn\: \mathrm{in}\: synapses$}{
        \If{$syn.from\: \mathrm{in}\: \EnabledBits{enc}\: \mathrm{AND}\: \IsConnected{syn}$}{
            \IncrementSignalCount{syn.to}
            }
    }
    $active \gets$ \GetCellsWithMostSignals{40}\;
    \;
    \tcc{Reinforce to ensure similar output on similar input}
    \ForEach{$syn\: \mathrm{in}\: synapses$}{
        \If{$syn.from\: \mathrm{in}\: \EnabledBits{enc}$}{
            \eIf{$syn.to \: \mathrm{in}\: active$}{
                \Grow{syn} \tcp{May connect disconnected synapse}
            }{
                \Shrink{syn} \tcp{May disconnect connected synapse}
            }
            }
    }
    \Return{active}\;
    \caption{The spatial pooling process on receiving an input} \label{alg:sp}
\end{algorithm}

\section{HTMRL}\label{sec:htmrl}
\begin{figure}[t]
    \includegraphics[width=\textwidth]{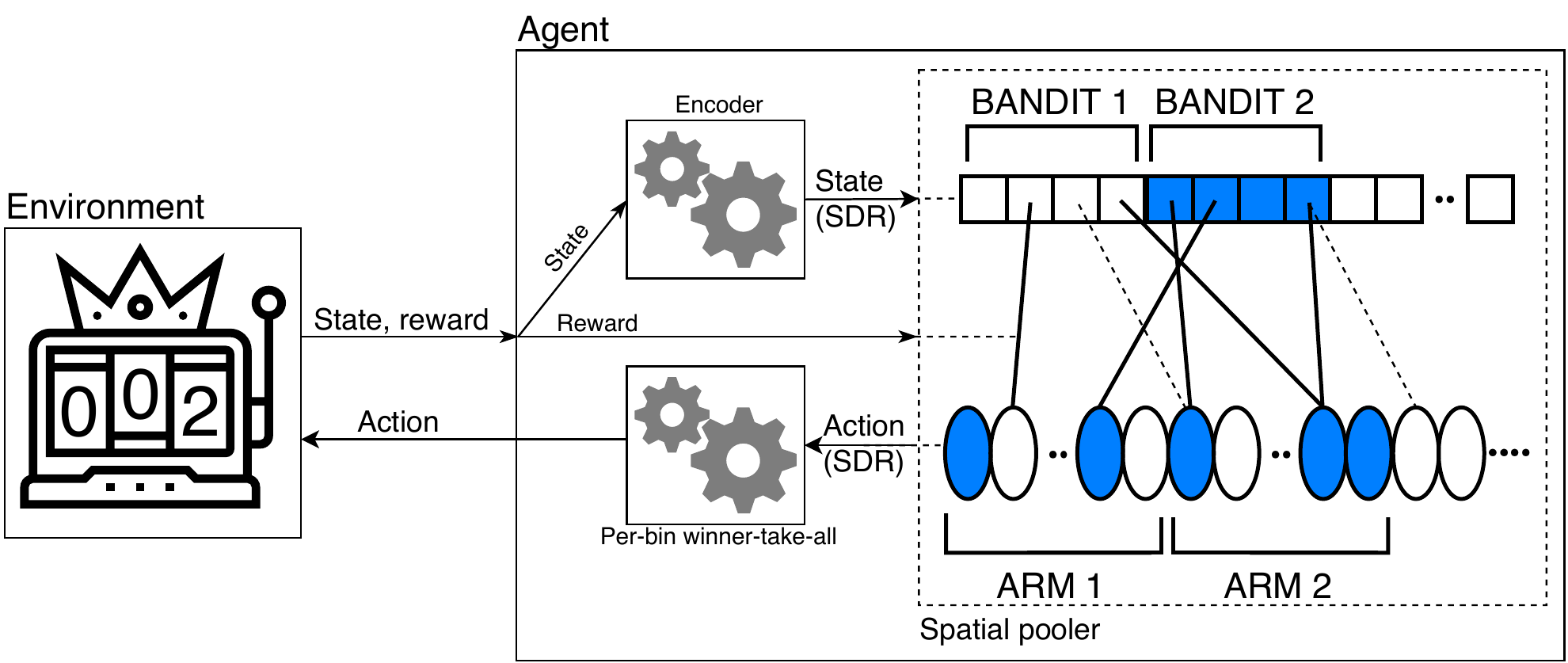}
    \caption{Overview of the HTMRL system, with coloured bits/cells being enabled/active, illustrated with a contextual bandit environment where the state is the bandit's index. The system receives a state/observation and runs this through an encoder that outputs an \gls{SDR} representing the state. Enabled bits send signals over their connected synapses to the spatial pooler cells, where the 40 cells receiving most signals activate, and the action with the most active cells is selected. Synapses from enabled bits to active cells of the chosen action then grow/shrink depending on the reward, and may become (dis)connected. Not all bits, cells and synapses are shown.
    } \label{fig:htmrl}
\end{figure}

The principle of \gls{RL} is considered to be one of the core components of human learning~\cite{mnih2015human,schultz1997neural}. As \gls{HTM} models the part of the brain that sets humans (and other mammals) apart from other animals, it should be possible to implement an \gls{RL} algorithm using the \gls{HTM} framework. In this section, we propose HTMRL; an \gls{RL} algorithm using only \gls{HTM}'s spatial pooler, with minimal modifications. We deliberately do not introduce any other elements (such as regular neural network layers) as one of our main goals is to show the feasibility of a strictly \gls{HTM}-based \gls{RL} algorithm. Figure \ref{fig:htmrl} summarises the algorithm.

Before presenting states or observations to the spatial pooler, they must be encoded as \glspl{SDR}. We can simply use existing numeric encoders~\cite{purdy2016encoding} here. Converting the spatial pooler's output to an \gls{RL} agent's output will require more effort however. The spatial pooler generates a selection of 40 out of 2048 cells, while an agent's policy returns the action to take. %
Assuming a deterministic policy, the policy in essence selects 1 action from the action space. With a finite action space of at most 2048 actions, it becomes relatively straightforward to map the spatial pooler output to action selection. With $|A|$ actions, we subdivide the 2048 cells into $|A|$ equally sized \textit{bins}, each representing one action. Any remaining cells are simply disabled. Every step, each bin is assigned a score equal to the number of selected cells it contains, and the highest scoring bin represents the chosen action. Ties are broken randomly. Note that the size of each bin should be at least 2 to avoid having 40-way ties at every step. Without changing the spatial pooler's architecture, this limits system to 1024 actions.

Finally, we need some method of introducing the reward signal to encourage good behaviour. First note that the internal synaptic structure of the spatial pooler remains unchanged: from each cell, synapses are grown to a randomly selected subset of input bits. In regular \gls{HTM} theory, all synapses between an active input bit and selected output cell are positively reinforced (i.e., strengthened) while those between active input bits and non-selected output cells are negatively reinforced (i.e., weakened)\footnote{To avoid confusion with \textit{reinforcing} in an \gls{RL} context, we will rely on the terms \textit{strengthen} and \textit{weaken} exclusively in the synaptic context.}. This behaviour is not desirable in an \gls{RL} context: cell selections leading to poor action selection should not be strengthened, and synapses to non-selected cells should not be weakened (as those may very well represent well-performing actions). As such, synapses to non-selected cells are \textit{not} modified, and the (by default positive) modification to synapses to selected cells are \textit{scaled with the reward}. 

This assumes that all positive rewards are desirable, while all negative rewards are not. In environments where this assumption does not hold, rewards should be normalised. A common normalisation method is to use each value's z-score by subtracting the mean ($\mu$) and dividing by the standard deviation ($\sigma$) of all values. As we do not know future rewards, we instead use the running mean and standard deviation. To avoid positively inflating rewards due to poor rewards at the start of training, we may also use a moving mean and standard deviation, over some window size $w$. The normalised reward at step $i$ is then defined as
\begin{equation}
r^{norm}_{i} = \frac{r_{i} - \mu_{w_{from}:w_{to}}}{\sigma_{w_{from}:w_{to}}}
\end{equation}
where $w_{to}$ is the current step, and $w_{from}$ is $w-1$ steps prior (or the first step, during the first $w-1$ steps).

The only core aspect of \gls{RL} still missing is the concept of \textit{exploration}, in which the learner is encouraged to take actions with uncertain results. The spatial pooler may learn to select decently performing actions, but miss strictly better performing actions entirely. Fortunately, a system achieving this goal is already implemented in the core \gls{HTM} theory. \textit{Boosting} artificially amplifies the incoming signal of rarely selected cells, meaning such a cell may be chosen over a more commonly selected cell, despite having fewer active incoming synapses. The \textit{boost strength} parameter then directly controls the degree of exploration.

Overall, HTMRL remains as biologically plausible as \gls{HTM}. The only change to its inner workings is the tweaked learning rule. We pose that it, just like the base rule, does not stray from the principles of Hebbian learning, generally considered to be biologically plausible~\cite{Mazzoni4433}. Specifically, not strengthening synapses to non-selected cells is equivalent to simply letting those cells deactivate before Hebbian learning is applied. Furthermore, Hebbian learning scaled with a reward signal has previously been proposed as biologically plausible behaviour~\cite{Legenstein8400}.

\section{Experiments \& Evaluation}\label{sec:results}
We apply HTMRL in several variations of a classic \gls{RL} environment to evaluate its performance. The goal of these experiments is twofold. We want to (1) investigate HTMRL's maximum capacity in terms of states and actions, and how well it scales with these and (2) compare its performance in changing environments to a simple, well-studied algorithm.
\subsection{HTMRL Capacity}
\subsubsection{Experiments}
We empirically evaluate how many states and actions HTMRL can support in a contextual bandit setting. Each bandit consists of $|A|$ arms, with a single \textit{winning} arm producing a reward of 1, and all others generating a negative reward of -1. The winning arm for each bandit is fixed during learning, and assigned at random during initialisation. The state $s = 1,...,|S|$ is simply the index of the current bandit, sampled uniformly at each step. The environment was designed to be very simple to learn: any failure to reach perfect performance in this environment should be the result of the HTMRL lacking sufficient representational capacity, as opposed to not being able to deduce complicated, stochastic mappings from state to expected reward. We experiment with values of 4, 16, 64, 256 and 1024 for either $|S|$ or $|A|$, while keeping the other fixed (at $|S|=20$ or $|A|=4$). We report the performance through a moving average of the 1000 most recent rewards at every step, and halt learning after selecting the winning arm for 100 subsequent steps, as this indicates that the algorithm has learned all correct actions. Experiments are repeated 20 times with different seeds, and the mean and standard deviation are reported. %

\begin{figure}[t]
    \centering
    \subfloat[Varying number of states\label{fig:states}]{
        \includegraphics[width=0.45\textwidth]{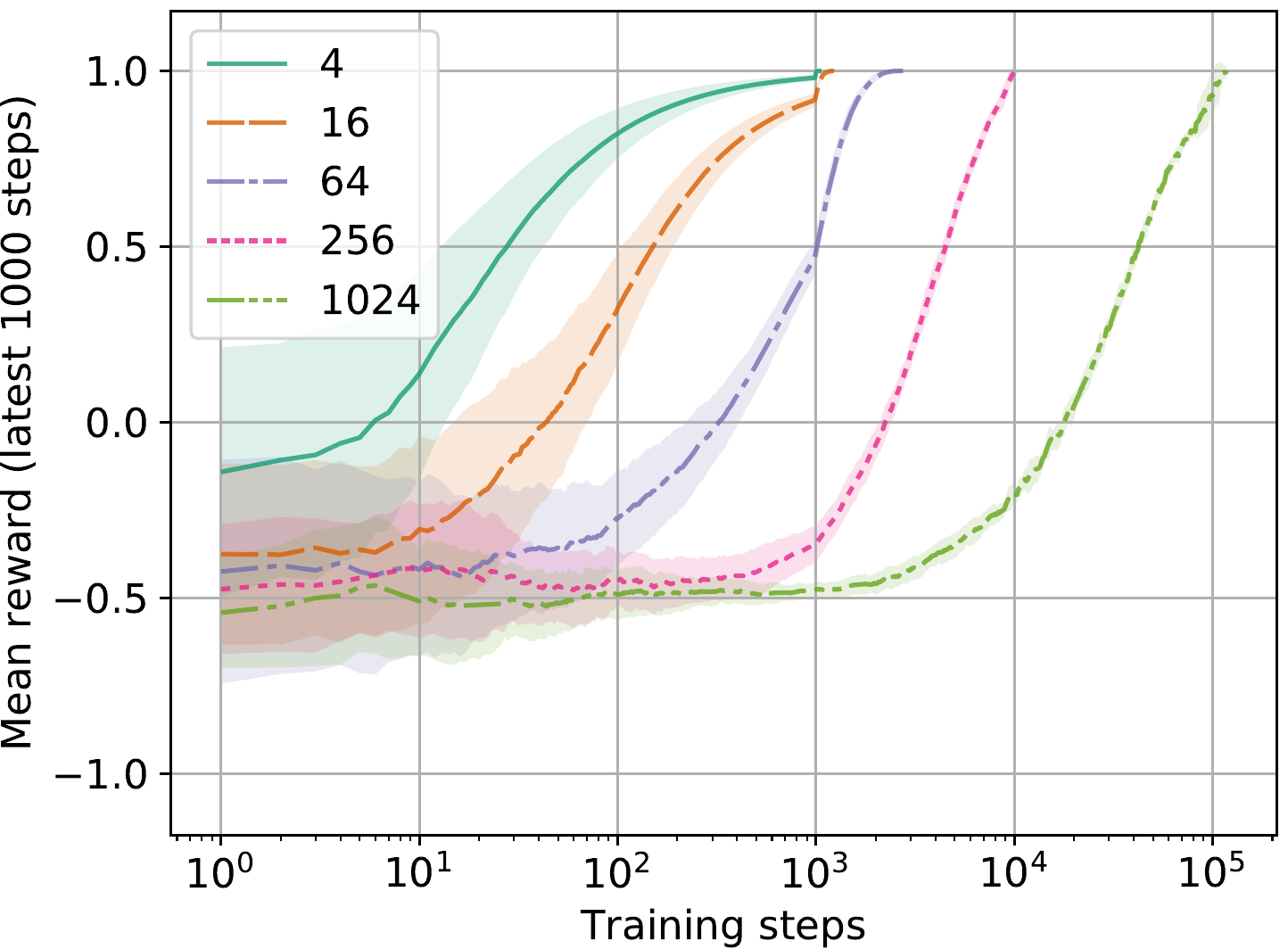}
    }
    \qquad
    \subfloat[Varying number of actions\label{fig:actions}]{
        \includegraphics[width=0.45\textwidth]{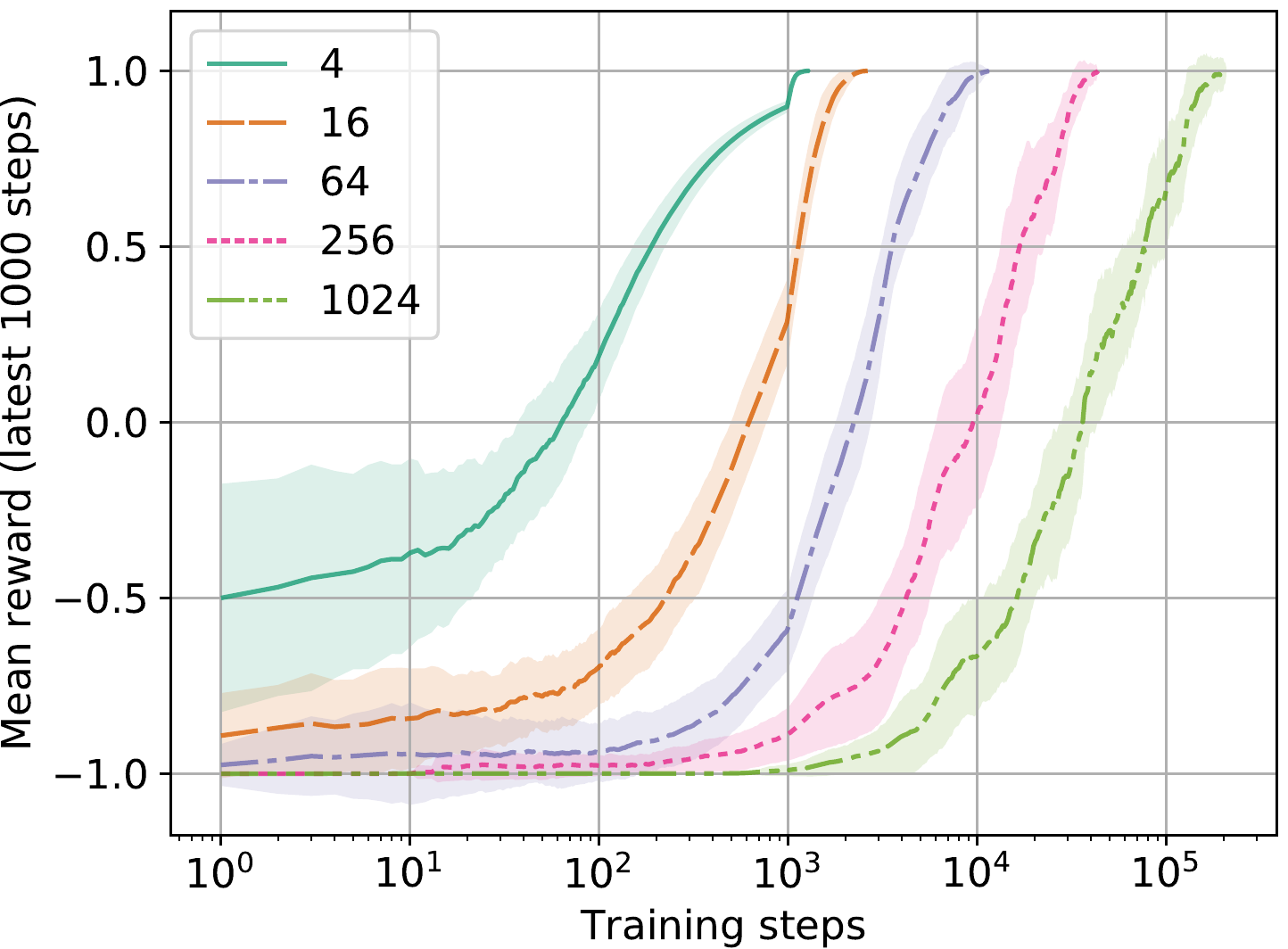}
        }
    \caption{HTMRL performance depending on the number of states or actions. Each reward value is the average reward over the last 1000 steps. The experiments were repeated 20 times, with lines and shaded areas representing the mean and standard deviation of those results, respectively. The line ends once optimal performance was achieved across all 20 repeats.}
    \end{figure}
\subsubsection{Results}
Figure \ref{fig:states} shows the results for state capacity. There is no theoretical limit on the number of states, as the input size of the spatial pooler is fully configurable. The network indeed succeeds in eventually memorising all optimal state-action combinations. There is however an obvious increase in training time when going from 256 to \SI{1024}{} states, with optimal performance being reached after over 10 times as many steps. At this point, the spatial pooler has an input size of \SI{20480}{} as we reserve 20 bits per state.
This increase in training time is likely due to a diminishing effect of the boosting system: as the time between visits to the same state increases, the effectiveness of giving priority to less frequently activated cells diminishes. There may be some merit to a boosting system per input bit, or a boosting window relative to $|S|$.

Next, Figure \ref{fig:actions} shows the results for the actions. Clearly, the network is eventually able to memorise all optimal actions. 1024 actions is the theoretical maximum capacity, with each action represented by only two cells. Without increasing the spatial pooler's size, HTMRL is thus able to support up to 1024 distinct actions, without requiring an infeasibly high number of training steps.
\subsubsection{Theoretical Analysis}
Training on 1024 bandits to cover all actions would be computationally infeasible with the current implementation. We instead statistically derive the probability that some combination of bandit index $i$ and optimal action $a$ could \textit{not} be represented by a (randomly initialised) HTMRL network modelling 1024 actions. A single synapse from each of the two cells representing $a$ to any of the input cells active for bandit $i$ would suffice to output the correct action. When each cell is initially connected to a fraction $c$ (by default $0.5$) of all input bits, and the encoded input \gls{SDR} consists of $n$ bits of which a fraction $d$ is active, the probability of action $a$ being attainable for input $i$ is\\
\begin{equation}
P^{(a,i)} = \left(1 - \frac{{n-dn \choose cn}}{{n \choose cn}}\right)^{2}
\end{equation} 

For a difficult but realistic case where $c=0.5, n=400, d=0.05$, the probability of the pair $(a,i)$ \textit{not} being representable is less than $\num{1.16e-6}$.
Furthermore, the probability of \textit{all} 1024 actions being attainable from a given state is over $0.9988$. If desired, these probabilities could be further ameliorated by increasing $\frac{d}{n}$, $c$ or the number of spatial pooler cells. Note that these probabilities ameliorate drastically if more than two cells represent an action, as it is then no longer required that \textit{all} of an action's cells are selected for it to be chosen.

\subsection{Non-Stationary Environments}
\subsubsection{Baseline Experiment}
We employ a more difficult variant of the bandit environment to evaluate HTMRL's ability to adapt to non-stationary environments, where sudden and drastic changes in the environment happen without being announced. In this environment, each arm is given a \textit{score}, sampled from a normal distribution ($\mu\myeq 0, \sigma\myeq 1$), and each time an arm is pulled, its actual reward is again sampled from a normal distribution ($\mu\myeq score, \sigma\myeq 1$). This makes deriving a policy significantly more difficult: pulling each arm once no longer suffices. The environment is changed every \SI{2000}{} steps, by completely reinitialising the bandit's arms. Knowledge learned up to that point becomes largely useless. To limit training time, the environment is limited to a single bandit, represented by 6 enabled bits. Each learner is given \SI{10000}{} steps, meaning there are 4 environment changes. As the reward function is stochastic, the experiment is repeated \SI{1000}{} times, and per-step rewards are averaged across all repeats, and averaged over a moving window of 10 steps within each repeat. As a baseline, we repeat the experiment with an $\epsilon$-greedy learner~\cite{sutton18}, which takes a random action with probability $\epsilon$, and the action thought to be optimal otherwise. We experiment with an $\epsilon$ of 0.1 and 0.01. 
\begin{figure}[t]
    \centering
    \subfloat[HTMRL vs. $\epsilon$-greedy\label{fig:nonstatic-htmrl}]{
    \includegraphics[width=0.45\textwidth]{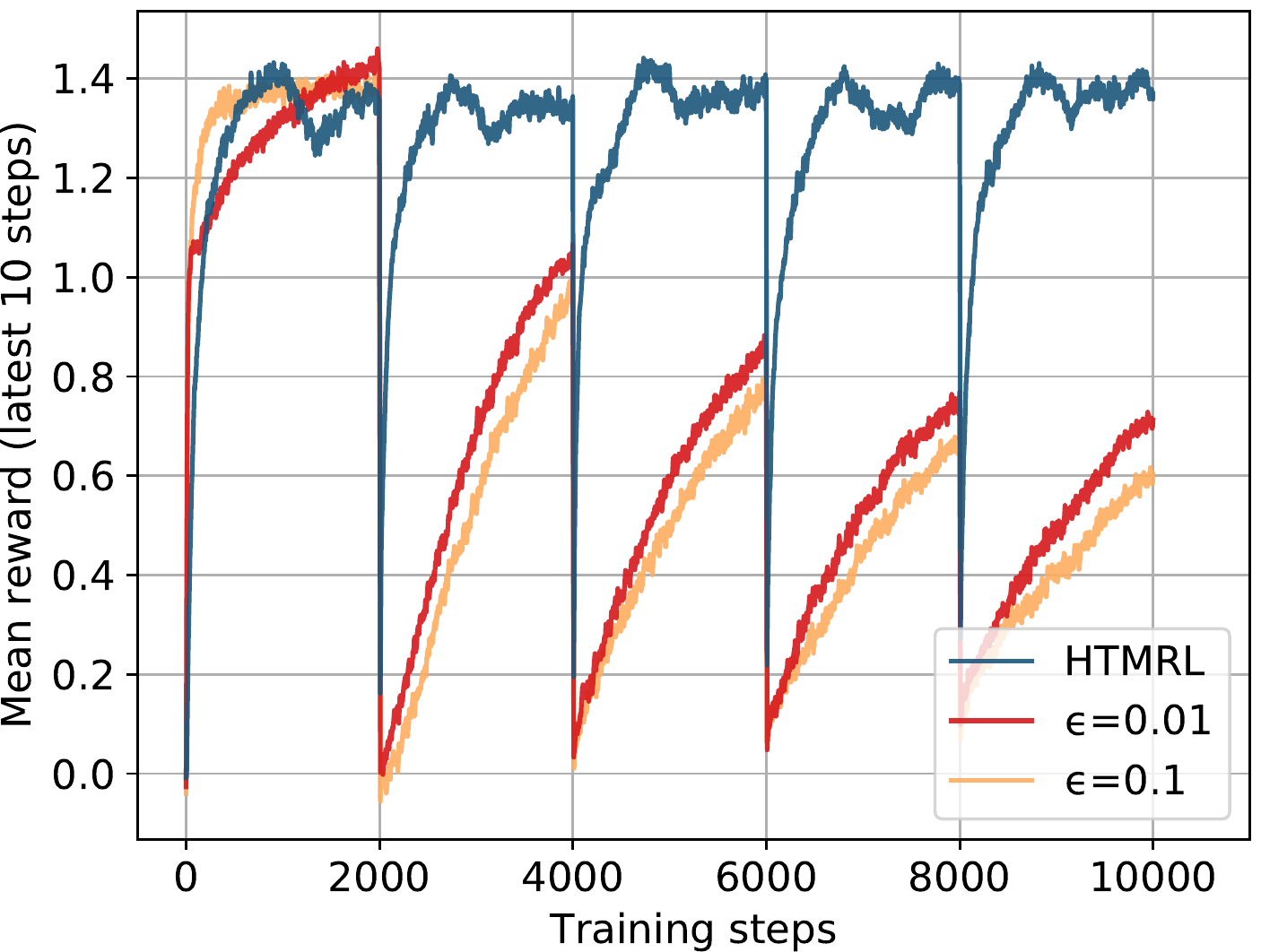}
    }
    \qquad
    \subfloat[HTMRL with restricted dimensions\label{fig:nonstatic-minhtmrl}]{
        \includegraphics[width=0.45\textwidth]{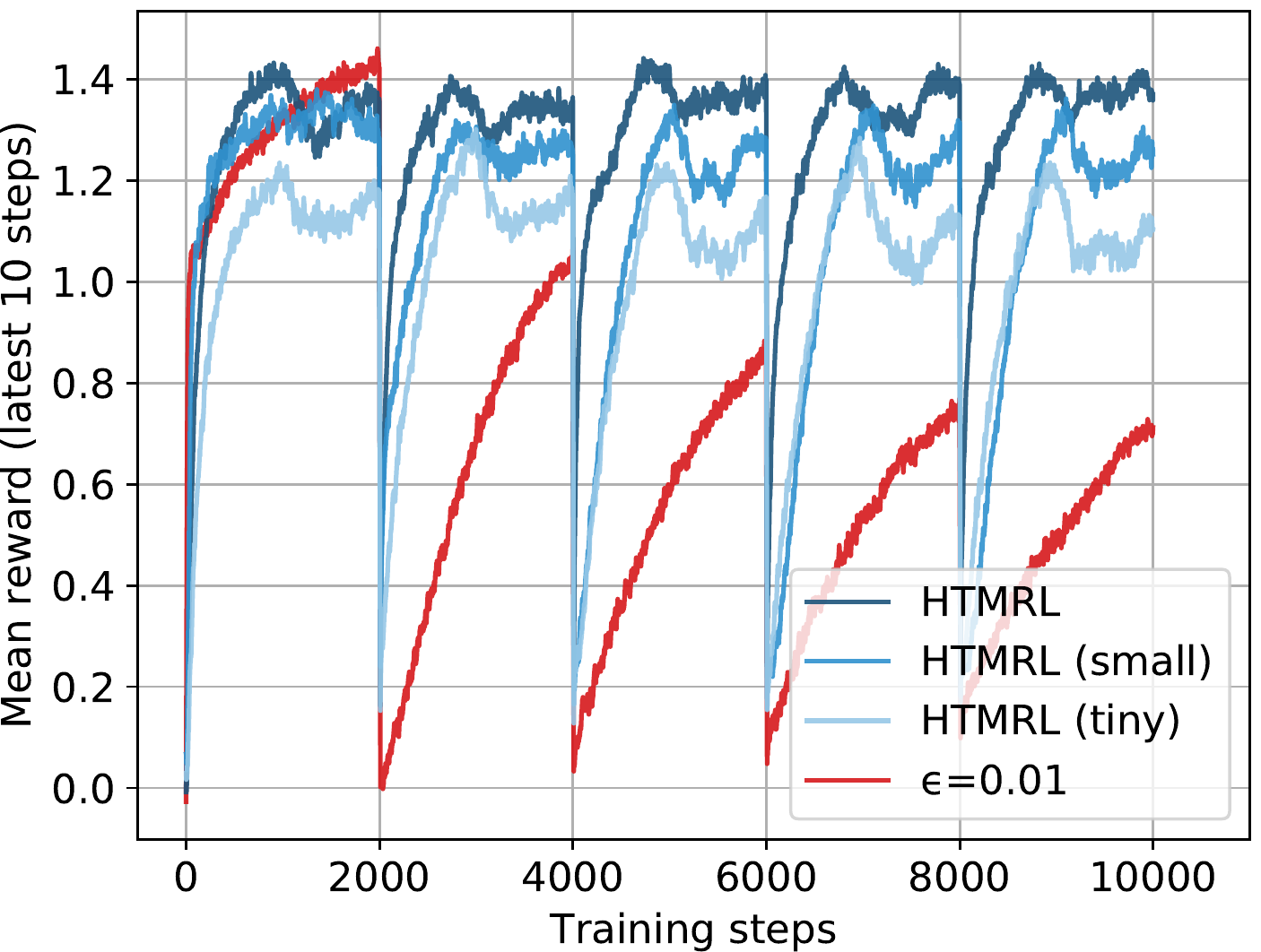}
    }
    \caption{Performance of HTMRL and $\epsilon$-greedy for a 10-armed stochastic bandit, reinitialising its arms every \SI{2000}{} steps.
    Even the \textit{tiny} HTMRL, over 100 times smaller than the default configuration, maintains its adaptive capability.
    }
    \end{figure}

Figure \ref{fig:nonstatic-htmrl} shows the resulting rewards. HTMRL adapts quickly to the reconfiguration, reaching its performance peak as fast as for the first initialisation. The $\epsilon$-greedy learner however suffers from the outdated knowledge, learning worse after every reconfiguration. This shows that \gls{HTM} can by default adapt quickly to changing patterns, as was previously demonstrated in sequence prediction applications~\cite{Cui2016Continuous,Struye2019Hierarchical}. 
\subsubsection{Restricted HTMRL} The baseline HTMRL is rather overdimensioned for this environment. It is possible that HTMRL simply employs a different subset of its many synapses to learn new patterns when the environment changes. We therefore repeat the experiment with severely restricted variations of the HTMRL network. The \textit{small} configuration maintains the baseline's 6 input bits, but restricts the cells from 2048 to 100, and the number of active cells from 40 to 10. The \textit{tiny} configuration is the smallest possible configuration for this environment, using only a single input bit and 20 cells, of which 2 activate. Figure \ref{fig:nonstatic-minhtmrl} shows the resulting performance compared to the full HTMRL system. Even the \textit{tiny} configuration maintains the ability to adapt quickly. Both the learning speed and maximum performance are lower with smaller networks, indicating the additional capacity is not wasted. The dip in performance that was noticeable with full-size HTMRL becomes considerably more pronounced with smaller networks. We investigate the cause below.
\begin{figure}[t]
    \centering
    \subfloat[Without boosting\label{fig:nonstatic-noboosthtmrl}]{
    \includegraphics[width=0.45\textwidth]{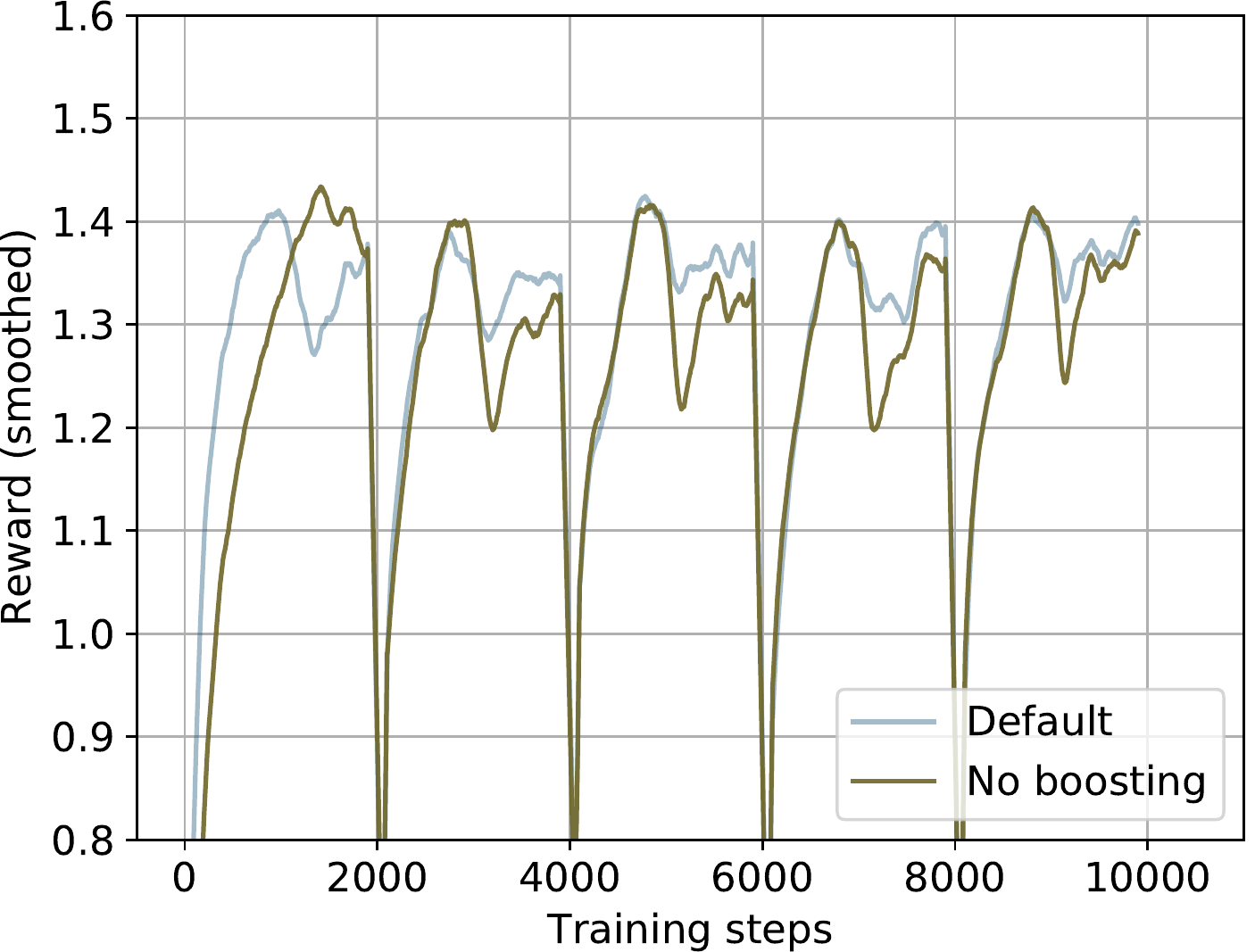}
    }
    \qquad
    \subfloat[Longer reward window\label{fig:nonstatic-longrewhtmrl}]{
    \includegraphics[width=0.45\textwidth]{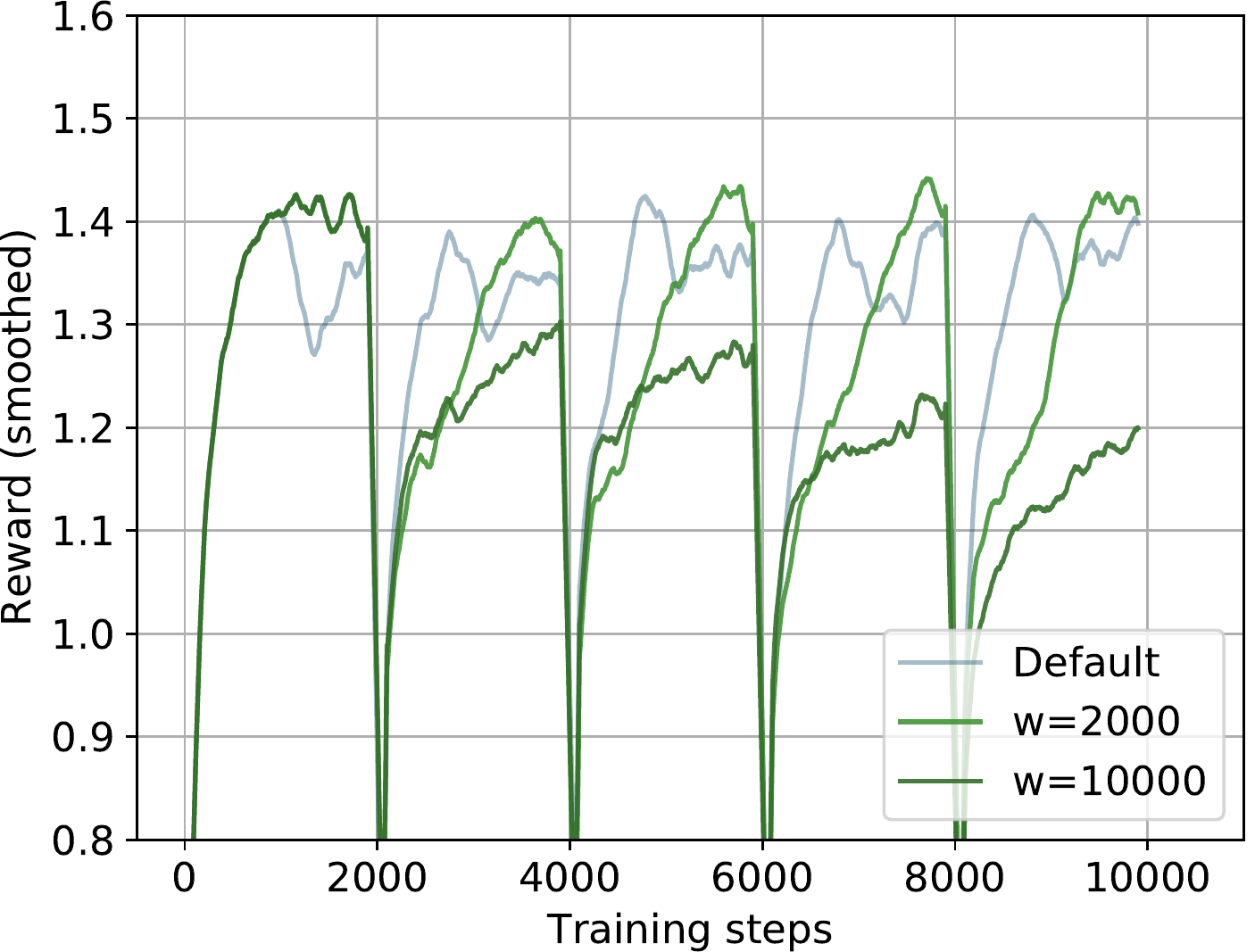}
    }\\
    \subfloat[Arm shuffling (HTMRL)\label{fig:nonstatic-reorderhtmrl}]{
    \includegraphics[width=0.45\textwidth]{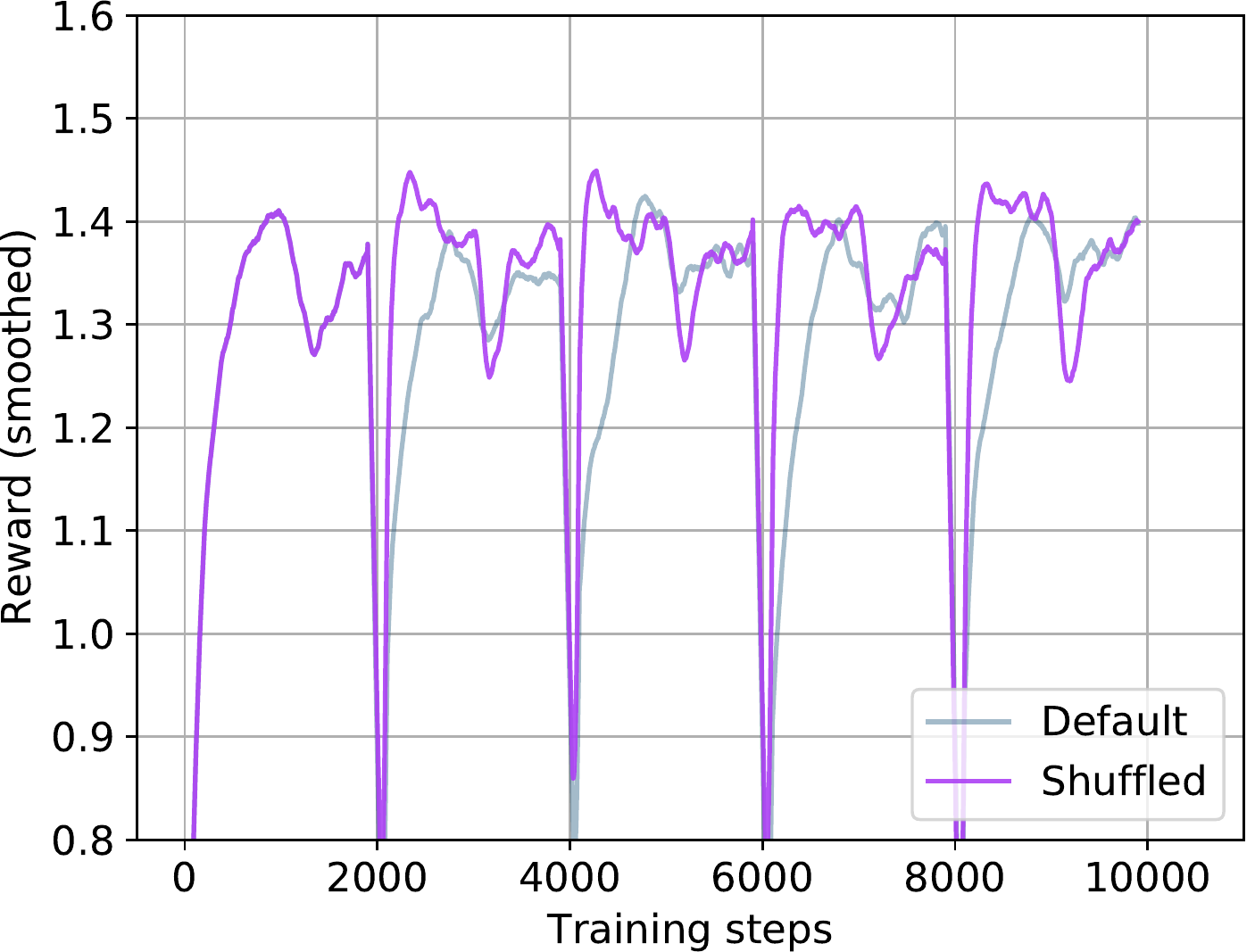}
    }
    \qquad
    \subfloat[Arm shuffling ($\epsilon$-greedy)\label{fig:nonstatic-reordereps}]{
    \includegraphics[width=0.45\textwidth]{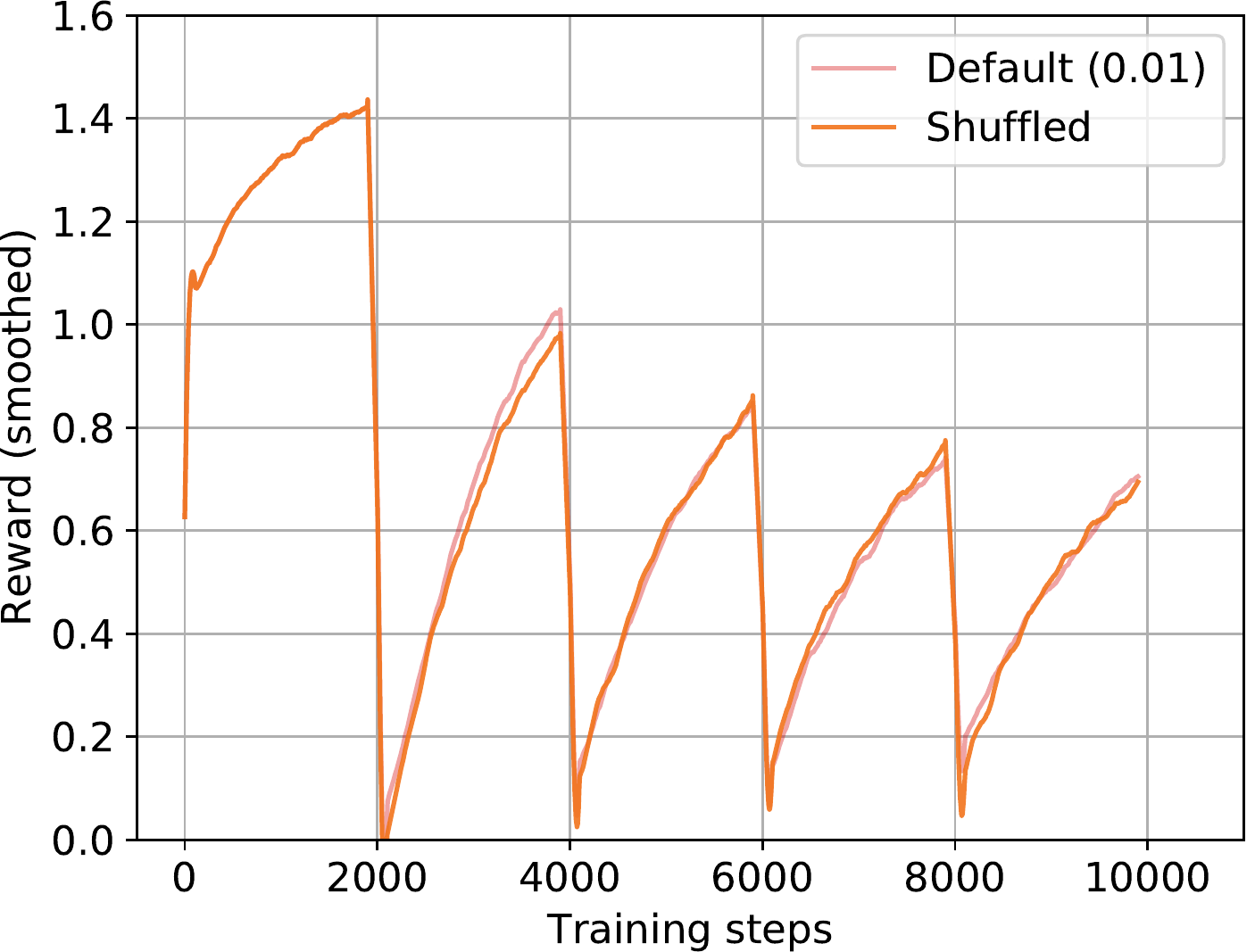}
    }
    \caption{Boosting and the reward normalisation window have a clear effect on HTMRL's learning speed and stability (top). When only shuffling the bandit's arms instead of reinitialising them, HTMRL can leverage previous knowledge while $\epsilon$-greedy cannot (bottom). To make overall trends clearer, the rewards were smoothed using a Savitzky-Golay filter.
    }
    \end{figure}
 \subsubsection{Boosting and reward normalisation} We now investigate the cause of the performance dip halfway between two environment changes. Boosting could be responsible by forcing the network to focus on rarely used cells, representing poor actions. Figure \ref{fig:nonstatic-noboosthtmrl} however shows that the effect only becomes more noticeable without boosting. Furthermore, initial learning becomes significantly slower.
 
 Next, we investigate the reward normalisation window $w$, by default \SI{1000}{}. Figure \ref{fig:nonstatic-longrewhtmrl} shows that when this window size is doubled, the effect disappears, but learning a new initialisation is slower. When normalising over the full history, the effect disappears entirely, but performance decreases after every reinitialisation. With higher $w$, the near-optimal rewards of the previous initialisations are remembered for longer, meaning good but non-optimal rewards for following initialisations are normalised to smaller values, slowing the learning process.
 
 The dip thus occurs when HTMRL has been pulling the best arm for at least \SI{1000}{} steps, as half of the rewards from the optimal arm are then normalised to negative values, shrinking the synapses that triggered the action. HTMRL will then select suboptimal actions in search of higher reward, but eventually returns to always selecting the optimal action as the mean reward reduces. The effect is more noticeable with smaller networks, as fewer synapses connect to the optimal action. Once those \textit{optimal action synapses} shrink, the only remaining synapses lead to suboptimal actions, while with a large network other, previously unused optimal action synapses remain. This makes full-size HTMRL with a large window more stable. The effect is however not necessarily undesirable; it in fact encourages the agent to explore when rewards no longer improve, and may be helpful in escaping a locally optimal policy.
\subsubsection{Meta-RL}
As a final experiment, we shuffle the bandit's existing arms, instead of completely reinitialising them. Figure \ref{fig:nonstatic-reorderhtmrl} shows that HTMRL now adapts quicker to the changes, while Figure \ref{fig:nonstatic-reordereps} shows that it makes no difference for the $\epsilon$-greedy learner. HTMRL is able to leverage knowledge of the previous ordering to learn the new configuration three times as fast. Knowledge of the arms' distributions remains relevant, and the learner minimally has to discover how these distributions were reassigned. The improved learning speed shows that HTMRL is able to take this approach. As such, HTMRL inherently performs well as a \gls{Meta-RL} algorithm.
\section{Related Work}\label{sec:rw}
Over the years, a few attempts to implement \gls{RL} using \gls{HTM} have been made. A first attempt is based on Hawkins' \gls{MPF}~\cite{hawkins2004intelligence}, of which \gls{HTM} is considered to be an implementation~\cite{rawlinson2012generating}. This approach uses \textit{sensorimotor} inputs (containing both states and actions) and predicts future inputs. Instead of incorporating the reward signal in the \gls{MPF}, it is introduced by artificially changing the layers' outputs to favour predictions containing states associated with high reward. It takes the agent over \SI{60000}{} iterations to achieve perfect play in rock-paper-scissors against an opponent rotating through the three actions. 
 Another approach performs robot control using only \gls{HTM}~\cite{Mai2013Simple}. Despite presenting an \gls{RL}-like problem, the algorithm is implemented as supervised learning, linking (visual) states to their optimal actions through classification, requiring training data. A final approach attempts to design a model-based \gls{RL} algorithm to play Atari games~\cite{quinonez2016model}. The transition function is modelled by an \gls{HTM} learning to predict future states, while the reward function is learned through a conventional neural network. This approach achieved moderate success in the game \textit{Breakout}, reaching 12 points, compared to an average of $0.7$ points using a random policy. In \textit{Pong} it however failed to outperform the random policy. Overall, no \gls{RL} algorithms using only \gls{HTM} have been published\footnote{A series of blog posts did claim to have built a functional \gls{HTM}-only \gls{RL} system, however we were unable to achieve better-than-random performance using the code provided. See: at \url{https://cireneikual.com/2014/11/09/my-attempt-at-outperforming-deepminds-atari-results-update-11/}}.

 Recently, considerable progress has also been made in the field of \gls{Meta-RL}. As state-of-the-art \gls{RL} algorithms such as Proximal Policy Optimization~\cite{Schulman2017Proximal} and Soft Actor-Critic~\cite{Haarnoja2018Soft} do not inherently handle non-stationary environments well, an additional algorithm is needed to meta-learn. Recent such approaches include Model-Agnostic Meta-Learning~\cite{Finn2017Model} and Proximal Meta-Policy Search~\cite{rothfuss2018promp}. In contrast, HTMRL adapts quickly to new variations of an environment, making it a promising \gls{Meta-RL} approach.
\section{Conclusions}\label{sec:conclusions}
In this paper, we presented HTMRL, the first strictly \gls{HTM}-based system not augmented with any other architectures capable of solving non-trivial \gls{RL} problems. With an action capacity of over \SI{1000}{} actions and a state capacity likely only limited by runtime of the implementation, the learner is not limited to very small environments only. One experiment showed that a core trait of \gls{HTM}, its ability to adapt quickly to changing patterns, does carry over to HTMRL. The learner was able to leverage previous knowledge when adapting to a 10-armed bandit shuffling its arms, reaching near-optimal performance three times as fast as when arms were completely reinitialised. As such, these results not only strengthen the value of the \gls{HTM} system, demonstrating its versatility, but also show that HTMRL is a viable \gls{Meta-RL} approach.

%
%
%

 \bibliographystyle{splncs04}
 \bibliography{bibliography}
\end{document}